\documentclass[runningheads]{llncs}
\usepackage[T1]{fontenc}
\usepackage{graphicx}

\usepackage{amsmath}
\usepackage{algorithm}
\usepackage{algpseudocode}
\usepackage{amssymb}
\usepackage{booktabs} 
\usepackage{multirow}
\usepackage{colortbl} 
\usepackage{marvosym}
\usepackage{ifsym}
\usepackage{bm}
\usepackage{hyperref}

\usepackage{tikz,xcolor,hyperref}

\definecolor{lime}{HTML}{A6CE39}
\DeclareRobustCommand{\orcidicon}{
\begin{tikzpicture}
\draw[lime, fill=lime] (0,0)
circle[radius=0.16]
node[white]{{\fontfamily{qag}\selectfont \tiny \.{I}D}};
\end{tikzpicture}
\hspace{-2mm}
}
\foreach \x in {A, ..., Z}{%
\expandafter\xdef\csname orcid\x\endcsname{\noexpand\href{https://orcid.org/\csname orcidauthor\x\endcsname}{\noexpand\orcidicon}}
}

\hypersetup{
    colorlinks=true,
    linkcolor=blue,
    citecolor=blue,
    urlcolor=blue
}
\usepackage{color}

\begin{document}

\title{Medical Image Segmentation via Single-Source Domain Generalization with Random Amplitude Spectrum Synthesis}
\titlerunning{RAS$^{4}$DG for Medical Image Segmentation}

\author{Qiang Qiao\inst{1,2} \hspace{-1.5mm}\orcidA{} \and
Wenyu Wang\inst{1,2} \hspace{-1.5mm}\orcidB{} \and
Meixia Qu\inst{1,2} \hspace{-1.5mm}\orcidC{} \and
Kun Su\inst{1,2} \hspace{-1.5mm}\orcidD{} \and \\
Bin Jiang\inst{1,2}\textsuperscript{(\Letter)} \hspace{-1.5mm}\orcidE{} \and
Qiang Guo\inst{3,4}\textsuperscript{(\Letter)} \hspace{-1.5mm}\orcidF{}}

%
\authorrunning{Qiao et al.}
%
\institute{School of Mechanical, Electrical \& Information Engineering,\\ Shandong University, Weihai, China\\
\email{jiangbin@sdu.edu.cn} \and
Shenzhen Research Institute of Shandong University, Shenzhen, China \and
School of Computer Science and Technology,\\ Shandong University of Finance and Economics, Jinan, China
\email{guoqiang@sdufe.edu.cn} \and
Shandong Provincial Key Laboratory of Digital Media Technology, Jinan, China}
\maketitle              
\begin{abstract}
The field of medical image segmentation is challenged by domain generalization (DG) due to domain shifts in clinical datasets.
The DG challenge is exacerbated by the scarcity of medical data and privacy concerns. Traditional single-source domain generalization (SSDG) methods primarily rely on stacking data augmentation techniques to minimize domain discrepancies.
In this paper, we propose Random Amplitude Spectrum Synthesis (RASS) as a training augmentation for medical images.
RASS enhances model generalization by simulating distribution changes from a frequency perspective. This strategy introduces variability by applying amplitude-dependent perturbations to ensure broad coverage of potential domain variations.
Furthermore, we propose random mask shuffle and reconstruction components, which can enhance the ability of the backbone to process structural information and increase resilience intra- and cross-domain changes. The proposed Random Amplitude Spectrum Synthesis for Single-Source Domain Generalization (RAS$^4$DG) is validated on 3D fetal brain images and 2D fundus photography, and achieves an improved DG segmentation performance compared to other SSDG models.
The source code is available at: \url{https://github.com/qintianjian-lab/RAS4DG}.

\end{abstract}

\section{Introduction}

Benefited by deep learning, medical image segmentation has witnessed considerable progress in the clinical practice of computer-aided diagnosis and treatment \cite{wangj}.
However, segmentation performance drops dramatically in the face of domain shifts such as different imaging modalities (e.g., CT and MR) or center data. To mitigate this, researchers explore domain adaptation (DA) and multi-source domain generalization (MSDG), which aim to transfer knowledge from annotated source domains to unlabeled target domains \cite{xiex}.

MSDG utilizes diversity from multiple source domains for generalization \cite{zhouk}. However, the success of data-driven approaches in the field of medical imaging is often hindered by the lack of sufficient data.
This is because 1) medical images are difficult to collect; and 2) manual labeling of fine-grained data from different domains is time-consuming and labor-intensive. Thus, training in an SSDG setting from only a single source domain faces serious challenges \cite{hus}. To deal with this problem, meta-learning \cite{lid} divides the source domain to simulate domain migration to unknown target domains \cite{liux}. In the context of medical image segmentation, SSDG is constrained by its reliance on data from a single domain, which is insufficient for capturing the varied distribution characteristics present across multiple domains. Recent studies have concentrated on enhancing the dataset to emulate the characteristics of target domains with uneven distributions \cite{ouyangc,zhangl,zhouz}. For example, RandConv \cite{xuz} leverages stochastic convolution as an innovative data augmentation method designed to maintain the integrity of shape and local texture information.
Unlike existing DG methods that standardize data to a uniform distribution, SAN-SAW \cite{pengd} encourages both intra-category compactness and inter-category separability. From the perspective of gradient information, SLAug \cite{suz} employs saliency-balancing to guide enhancement of global and local regions. In addition, generating adversarial samples is also employed to bolster the generalization capabilities of models \cite{volpir,hendrycksd}.

\begin{figure}[t]
\centering
\includegraphics[width=4.3in]{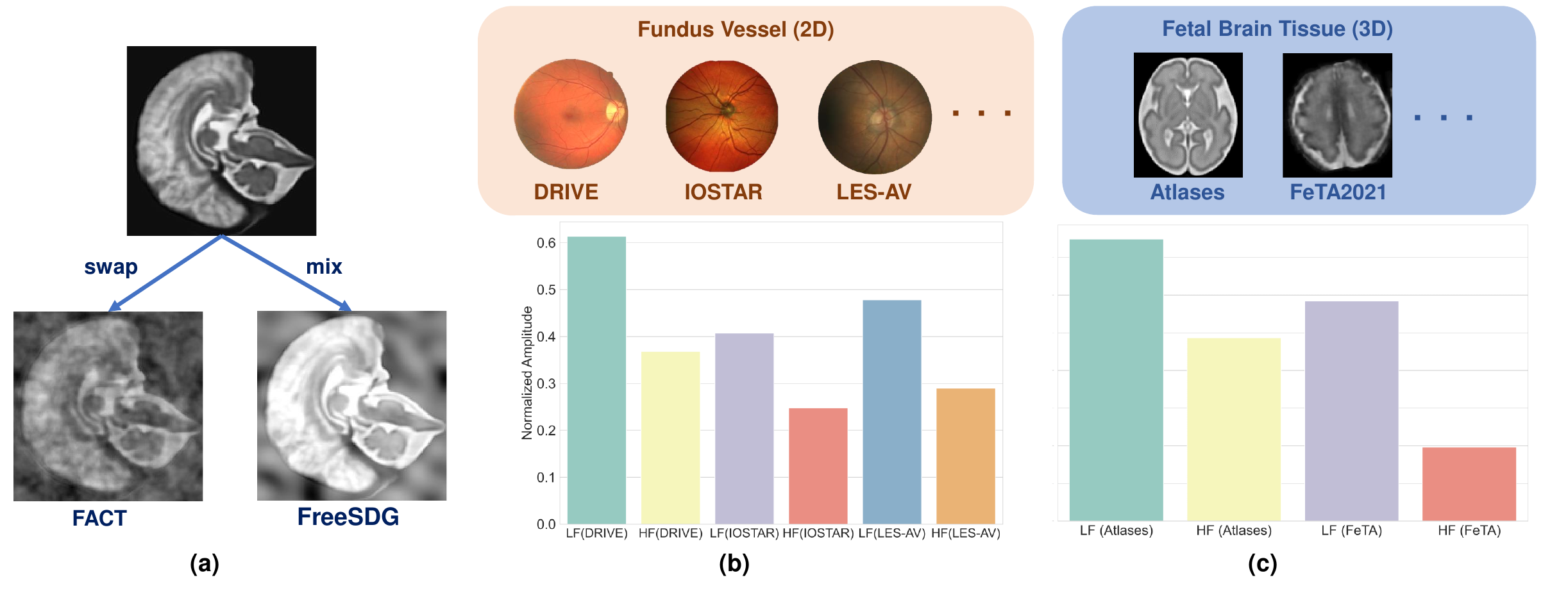}
\caption{Visualization of various results. (a) Results of different methods on the Atlases dataset.
(b) 2D datasets and (c) 3D datasets used in our experiments and the statistical analysis of the image amplitude for high frequency (HF) and low frequency (LF).
} \label{fig1}
\end{figure}

Analyzing medical images in the frequency domain provides a more profound understanding of the underlying patterns.
Specifically, the Fourier transform of images produces both amplitude and phase spectra \cite{liuq}.
We identify two key insights from this transform: the phase spectrum captures local details through contours, while the amplitude spectrum reflects low-level information such as texture.
There exist some works that focus on considering the amplitude spectrum in different images.
Chattopadhyay et.al \cite{cp} discovered that introducing variable perturbations to the amplitude information in natural images can effectively mitigate domain shifts.
The amplitude spectra of any two different images in a single source domain are swapped and mixed in FACT \cite{xuq} and FreeSDG \cite{lih}, respectively. When applying them in the Atlases dataset, we observe instabilities: FACT produces artifacts and FreeSDG causes significant color changes, as shown in Fig. \ref{fig1}(a).
Besides, we analyze 2D and 3D image datasets from various domains, and reveal that high-frequency bands of the amplitude spectrum show less variation than low-frequency bands (see Fig. \ref{fig1}(b) and (c)).

Based on these observations, we explore the frequency bands from the amplitude and propose a novel medical SSDG method named Random Amplitude Spectrum Synthesis for Single-Source Domain Generalization (RAS$^{4}$DG).
During the RASS process, high-frequency bands are subjected to more pronounced perturbations than low-frequency bands.
But images from different domains are heterogeneous.
It is difficult for the backbone network to capture structural information among images only by reducing the stylistic differences. Thus, we resort to make the backbone mine inter-image structural information to increase its context extraction capability. Specifically, local regions of the image are randomly masked with structural information, and the pixels within these regions are shuffled. To reduce feature redundancy and capture key information, a plug-and-play reconstruction convolution is further integrated into the backbone.


\section{Method}
\begin{figure}[t]
\centering
\includegraphics[width=4.3in]{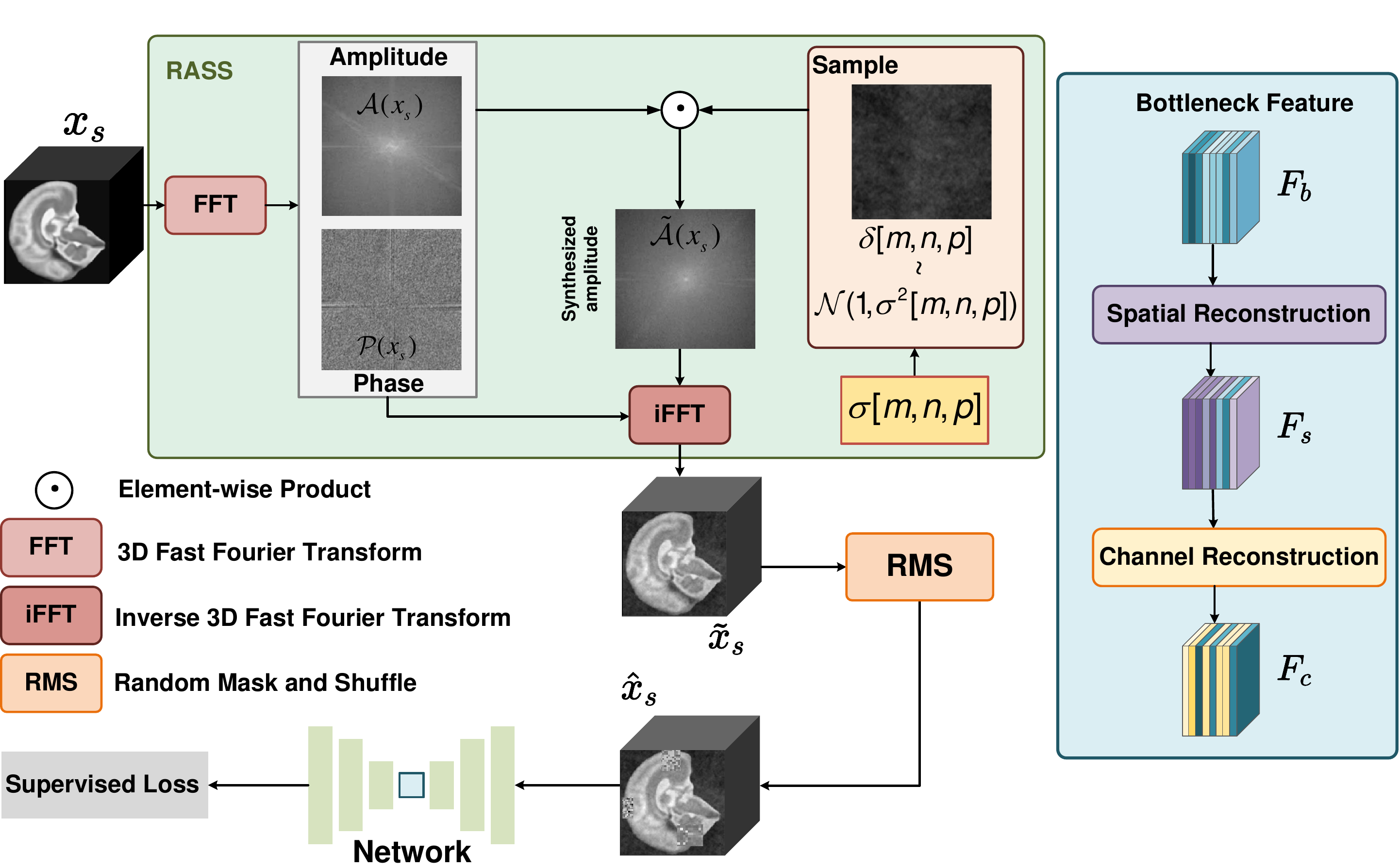}
\caption{Pipeline of our RAS$^{4}$DG.
The bottleneck of the network is shown in right side.
} \label{fig2}
\end{figure}

Let $\mathcal{S}$ and $\mathcal{T}$ represent source and target domains, respectively, which share the same label space. The training set $\mathcal{S}=\{(x_{s},y_{s})\}^{N}_{s=1}$ contains $N$ training pairs,
where $x_{s}$ is the $s$-th image from the source domain, and $y_{s}$ is the corresponding ground truth label.
The data from the target domain $\mathcal{T}$ is unknown and is not involved in the training process.
Our RAS$^{4}$DG framework is depicted in Fig. \ref{fig2}.
The amplitudes obtained by applying the Fourier transform on $x_{s}$ are randomly perturbed, and the inverse Fourier transform of the perturbed amplitudes and original phases is performed to recover the image $\tilde{x}_{s}$.
After performing the style transformation, structural information is interfered by using random mask and shuffle (RMS) to obtain $\hat{x}_{s}$.
Finally, the reconstruction design (RSD) at the bottleneck layer of the network eliminates redundant features.

\subsection{Random Amplitude Spectrum Synthesis}

From Fig. \ref{fig1} (b) and (c), we can observe that the variance of low-frequency bands of the amplitudes is larger than that of high-frequency bands in both 2D and 3D medical images. Meanwhile, there is an inconsistency in the variance of the high- and low-frequency bands of these amplitudes. Specifically, the high-frequency bands exhibit relative consistency across domains, whereas low-frequency bands display significant disparities. This suggests that high-frequency bands should be subjected to more variation than low-frequency bands, which help the model to extract domain-invariant features and prevent overfitting to a single domain. Thus, our method differentially perturbs the amplitude spectrum of images, with a particular focus on amplifying disturbances within the high-frequency range.

Given a 3D medical image $x_s$, we can extract its amplitude $\mathcal{A}(x_s)$ and phase $\mathcal{P}(x_s)$ by the Fourier transform. To minimize domain disparity and simulate potential changes in the unknown domain $\mathcal{T}$,
we inject perturbations into the amplitude spectrum of source domain images, and apply the perturbation magnitude $\sigma[m,n,p]$ uniformly across the frequency spectrum, where $m, n, p$ denote spatial frequencies.
Nonetheless, it is incapable of discerning the unique attributes of various frequency bands, which may lead to ignoring detailed information on high-frequency components vital for distinguishing domain differences.

To address the intrinsic complexity of medical images and exploit the nuanced differences between high- and low-frequency components, we propose Random Amplitude Spectrum Synthesis (RASS) to proportionally perturb the amplitude spectrum based on spatial frequency. Our perturbation function $g(\cdot)$ is then applied to the amplitude spectrum $\mathcal{A}(x_s)$ to achieve the desired perturbation. For each spatial frequency $[m,n,p]$, we introduce a random element $\delta[m,n,p]$ sampled from a Gaussian distribution (i.e., $\delta[m,n,p] \sim \mathcal{N}(1,\sigma^{2}[m,n,p])$) to modulate the spectral components, which can be formalized as:
\begin{equation}
\label{eq1}
g(\mathcal{A}(x_s))[m,n,p] = \delta[m,n,p]\mathcal{A}(x_s)[m,n,p].
\end{equation}

This frequency-dependent perturbation magnitude $\sigma[m,n,p]$ varies with frequency and allows a controlled increase in the perturbation amplitude of the high-frequency component.
It is tailored to emphasize the high-frequency content that is often more susceptible to variations in medical imaging environments and devices.
The magnitude of perturbation for spatial frequency is defined as:
\begin{equation}
\label{eq2}
\sigma[m,n,p] = \left(2\alpha \sqrt{\frac{m^2 + n^2 + p^2 }{H^2 + W^2 + D^2}}\right)^\gamma + \beta,
\end{equation}
where $\alpha$ controls the overall amplitude of the perturbation, $\gamma$ determines the rate at which the perturbation increases with frequency, and $\beta$ ensures that a minimal level of perturbation is uniformly applied, providing a baseline for all frequencies. This polynomial function ensures dynamic interactions within the amplitude spectra.
In other words, higher frequencies are subject to a more pronounced perturbation in comparison to lower frequencies.
The workflow of the RASS is summarized in Fig. \ref{fig2}.
After obtaining the randomly synthesized amplitude, we combines it and its original phase to recover the image $\tilde{x}_{s}$. Fine amplitude processing can simulate more realistic domain-varying medical images.
Thus, RASS enhances the ability of model to capture domain-invariant features in the image domain by applying different levels of frequency perturbation.

\subsection{Random Mask Shuffle and Reconstruction}

RASS has been presented to address with artifacts and unknown domain distributions in this paper.
However, current SSDG methods still face limitations in learning from heterogeneous structures.
PatchShuffle \cite{kangg} shows that random patch shuffling is more robust to noise and local image changes.
To handle local variations present in medical images and inject randomness, we employ a strategy of randomly masking image regions and shuffling the pixels within these sections.
This RMS strategy not only addresses the issue of local discrepancies but also compels our model to derive more robust feature representations by embracing randomness.
Specifically, random positions and sizes are determined from the image $\tilde{x}_{s}$ after RASS, which aims to select the regions to be masked in the image. Then pixels within each designated region are shuffled and filled to their original locations. This pixels shuffling enables weights to be shared between neighboring pixels, which increases the complexity of the backbone. This is useful for higher level feature mapping. At last, we obtain the image $\hat{x}_{s}$. The injection of stochasticity allows our RASS and RMS strategies to enhance the adaptability of the network to various unforeseen changes in medical images.

\begin{figure}[t]
\centering
\includegraphics[width=4.3in]{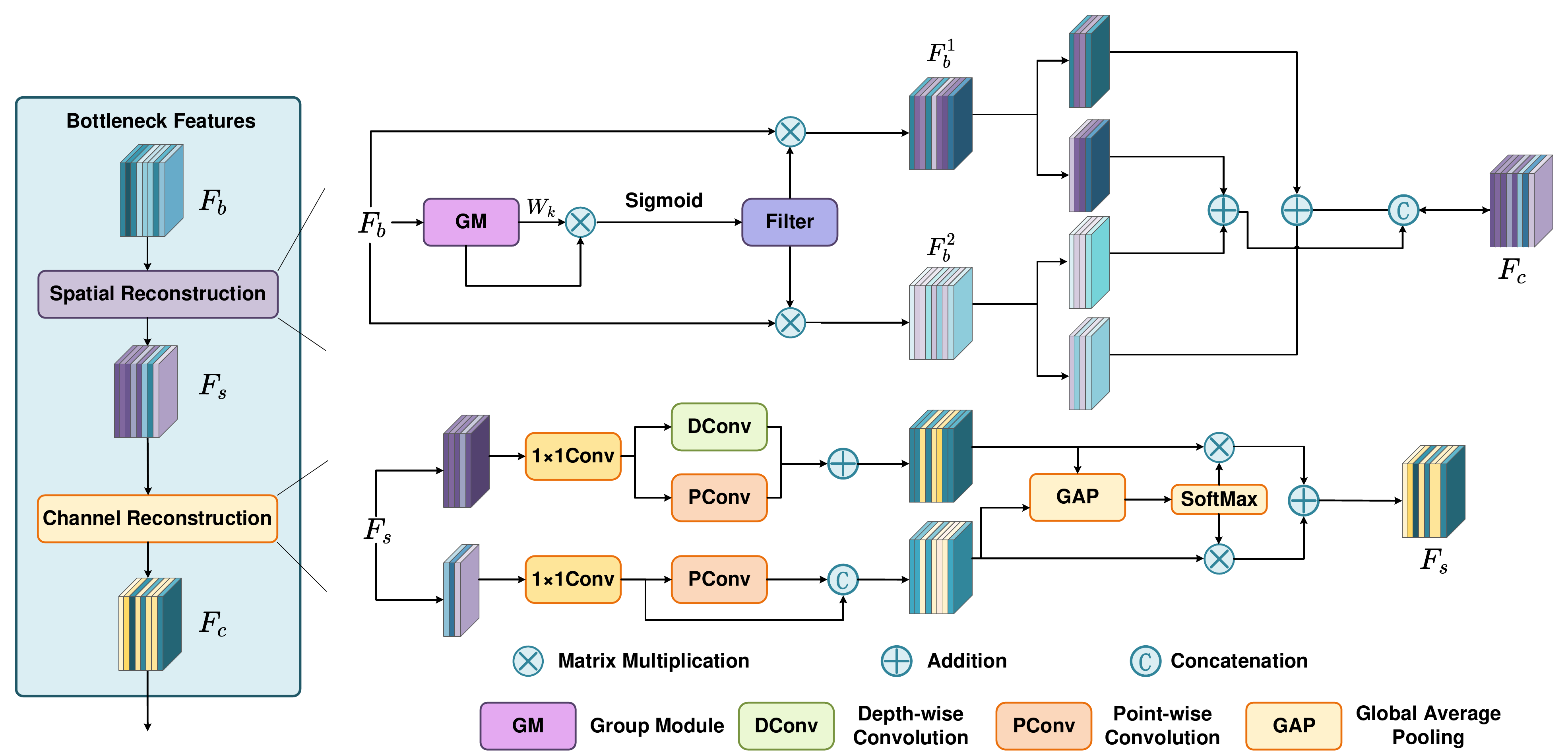}
\caption{Detail view of RSD. $F_{b}$ is reconstructed from spatial and channel dimensions.} \label{fig3}
\end{figure}

Besides, feature redundancy hampers effective feature learning and diminishes the capacity of the network for feature representation. Thus, a plug-and-play feature reconstruction design tailored for segmentation models is proposed.
RSD is used to enhance the capture of stylistic information after RASS and the feature representation within the region shuffled after RMS. We further optimize the features from the spatial and channel dimensions, as detailed in Fig. \ref{fig3}.

For the spatial dimension, we employ a strategy of `separation and reconstruction' for the bottleneck layer features $F_{b}$, where we utilize trainable scaling factors from group normalization to evaluate the significance of individual features and measure the variability in spatial information. The features $F_{b}$ are categorized into high-information $F^{1}_{b}$ and low-information $F^{2}_{b}$ based on weights and cross-fused. This nuanced differentiation of features based on their information content allows for strategic recombination of a diverse set of features. This not only preserves the feature space but also enriches the representational power of the features. Furthermore, to refine and eliminate redundant features, we implement a strategy of `divide-transformation-merging' along the channel dimension. Specifically, channels are divided, with one part enriched via depth-wise and point-wise convolutions, and the other part supplemented with point-wise convolutions.
The enriched and supplemented features are merged by using the SKNet \cite{lix}.
This process merges the richly learned features with supplementary ones to refine the channel features.
By merging these features, we enhance the ability of our network to capture complex patterns and nuances within medical images, which is crucial for accurate segmentation.

\setcounter{footnote}{0}

\section{Experiments and Results}
\subsubsection{Dataset.}
To demonstrate the versatility of RAS$^{4}$DG, we evaluated it using 3D and 2D medical image segmentation tasks.
For the 3D datasets, the source domain dataset of fetal brain tissue is obtained from Atlases, which contains 47 samples \cite{fidonl,gholipoura,wuj}.
The target domain dataset is composed of 80 MRI volumes from the Fetal Brain Tissue Annotation and Segmentation Challenge (FeTA) 2021 dataset \cite{payettek}.
In addition to the background, both the source and target domains included seven distinct categories: external cerebrospinal fluid (eCSF), grey matter (GM), white matter (WM), deep grey matter (dGM), cerebellum (CBM), lateral ventricles (LV), and brainstem (BS).
We further integrated 2D fundus vessel images from three different public datasets. The source domain images are derived from the DRIVE \cite{drive} fundus vascular dataset, which includes images of 20 patients and annotations.
 To assess the generalization capabilities of RAS$^{4}$DG, we utilized the IOSTAR \cite{iostar} and LES-AV \cite{lesav} datasets as target domains, containing  images from 30 and 22 patients, respectively.

\begin{figure}[t]
\centering
\includegraphics[width=\textwidth]{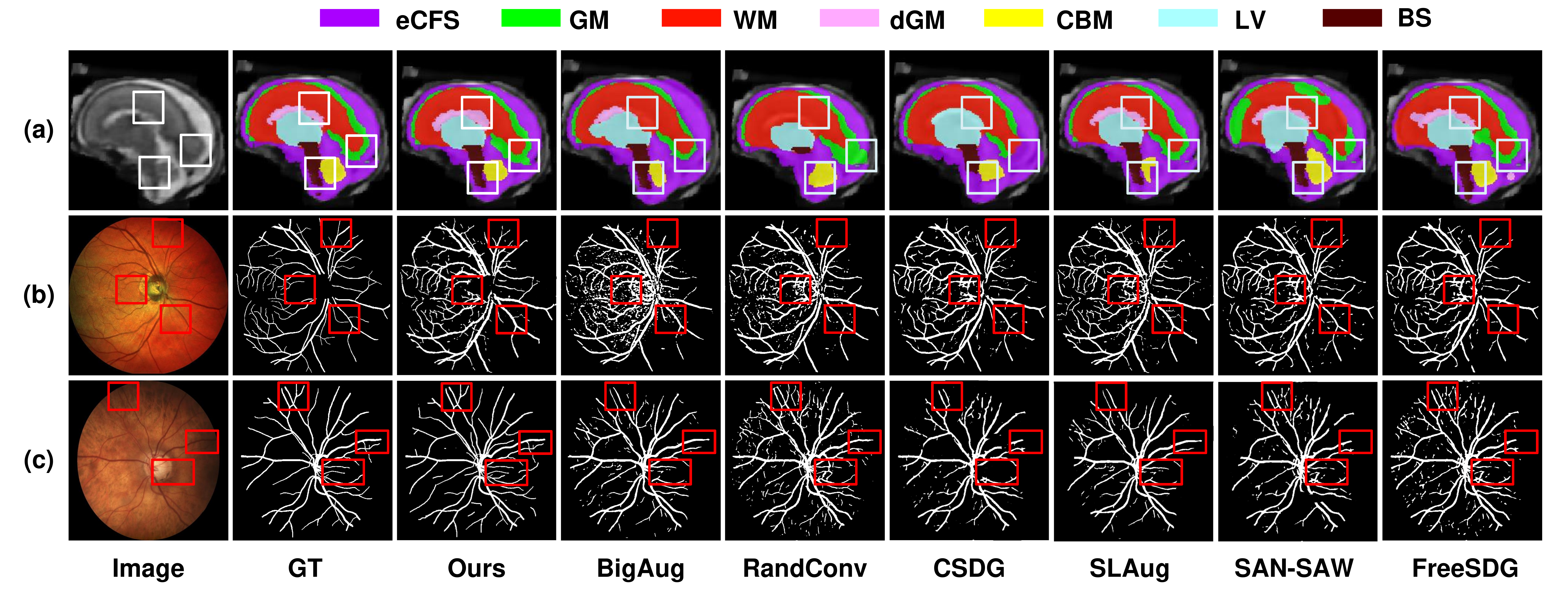}
\caption{Visualization of segmentation masks predicted by different methods in different datasets. (a) FeTA2021, (b) IOSTAR, and (c) LES-AV.} \label{fig4}
\end{figure}

\subsubsection{Implementation Details.}
All experiments are trained on NVIDIA A100 GPU (80G Memory) using the PyTorch framework. The results are evaluated using the Dice similarity coefficient (DSC) scores. We use SegResNet \cite{myronenkoa} as the backbone model. The training epoch is set to 300, and the initial learning rate is set to 0.0001. The network parameters is handled by the Adam optimizer. Furthermore, we implement a `Ploy' learning strategy to dynamically adjust the learning rate. All 3D images are resized to $144 \times 144 \times 144$ voxels, while 2D images are resized to $512 \times 512$ pixels. Our RAS$^{4}$DG is trained from scratch with early stopping. In RASS, hyperparameters $\alpha$, $\beta$, and $\gamma$ are set to $3.0$, $0.25$, and $2.0$, respectively. More details can be found in the supplementary materials.

\subsubsection{Comparison with the State-of-the-arts.}
To demonstrate the effectiveness of the proposed RAS$^{4}$DG, we report its average performance against state-of-the-art (SOTA) methods across three training rounds on 3D and 2D medical image datasets. We compare RAS$^{4}$DG against two benchmark methods: the `Intra-Domain' (an upper bound by training and testing within the same target domain) and the `w/o SSDG' (a lower bound by training in the source domain and testing in the target domain). We also explore two DA methods (CutMix \cite{yuns} and DSA \cite{hanx}), two Fourier-transform-based DG techniques (FedDG \cite{liuq} and FreeSDG \cite{lih}), and five SSDG methods (BigAug \cite{zhouz}, RandConv \cite{xuz}, CSDG \cite{ouyangc}, SLAug \cite{suz}, and SAN-SAW \cite{pengd}). All competitive methods are re-implemented based on their official repositories and utilize the same backbone network as our method.
The experimental results are listed in Table \ref{tab:table1}. Our method outperforms the two DA methods as well as the SSDG methods in terms of performance. Compared with the baseline model (w/o SSGD), our RAS$^{4}$DG achieves an average improvement of 8.15\% in DSC for the FeTA2021 dataset, 8.43\% for the IOSTAR dataset, and 13.71\% for the LES-AV dataset. Additionally, the proposed method narrows the DSC gap with supervised training to 4.80\% in FeTA2021, 8.38\% in IOSTAR, and 4.41\% in LES-AV. It is worth noting that FedDG is a Fourier based method for MSDG. The experiments on the 3D dataset do not fulfill the multi-source condition, so the results are not provided.
Fig. \ref{fig4} show some segmentation results. As evident in the boxed area, our RAS$^{4}$DG  enhances local details compared to other methods. Our results match the ground truth most accurately.

\begin{table}[t]
\centering
\caption{Comparisons with SOTA methods using DSC scores (\%). The three columns for FeTA2021 results are abnormal, normal, and average DSC. Best scores are in \textbf{bold}.}
\label{tab:table1}
\newcolumntype{P}[1]{>{\centering\arraybackslash}p{#1}}

\begin{tabular}{p{2.2cm}|P{1.7cm}P{1.7cm}P{1.7cm}|P{1.7cm}|P{1.7cm}}
\hline
\multirow{2}{*}{Methods} & \multicolumn{3}{c|}{3D}       & \multicolumn{2}{c}{2D}               \\ \cline{2-6}
                        & \multicolumn{3}{c|}{FeTA2021} & \multicolumn{1}{c|}{IOSTAR} & LES-AV \\ \hline
Intra-Domain            & $76.98_{\pm 0.17} $    & $85.74_{\pm 0.23}$    & $81.36_{\pm 0.21}$   & $74.24_{\pm 0.13}$                       & $77.29_{\pm 0.31}$  \\
CutMix \cite{yuns}                 & $74.10_{\pm 0.13}$    & $79.70_{\pm 0.50}$    & $76.20_{\pm 0.20}$   & $64.82_{\pm 0.31}$                       & $71.48_{\pm 0.21}$  \\
DSA \cite{hanx}                    & $73.44_{\pm 0.61}$    & $79.91_{\pm 0.42}$    & $75.93_{\pm 0.52}$   & $65.58_{\pm 0.19}$                       & $71.03_{\pm 0.11}$  \\
FedDG \cite{liuq}                  & -    & -    & -   & $65.30_{\pm 0.33}$                       & $71.33_{\pm 0.71}$  \\ \cline{1-6}
w/o SSDG                & $66.78_{\pm 0.14} $    & $70.04_{\pm 0.18}$    & $68.41_{\pm 0.14}$   & $57.43_{\pm 0.21}$                       & $59.17_{\pm 0.21}$  \\
BigAug \cite{zhouz}                 & $70.90_{\pm 0.41}$    & $79.27_{\pm 0.31}$    & $74.15_{\pm 0.26}$   & $62.43_{\pm 0.13}$                       & $66.87_{\pm 0.14}$  \\
RandConv \cite{xuz}               & $70.03_{\pm 0.21}$    & $76.81_{\pm 0.35}$    & $73.42_{\pm 0.22}$   & $62.01_{\pm 0.16}$                       & $67.98_{\pm 0.21}$  \\
CSDG \cite{ouyangc}     & $69.38_{\pm 0.21}$    & $74.88_{\pm 0.13}$    & $72.13_{\pm 0.15}$   & $61.94_{\pm 0.14}$                       & $68.30_{\pm 0.12}$  \\
SLAug \cite{suz}                 & $71.52_{\pm 0.13}$    & $77.04_{\pm 0.32}$    & $74.28_{\pm 0.27}$   & $62.36_{\pm 0.28}$                       & $68.79_{\pm 0.29}$  \\
SAN-SAW \cite{pengd}                & $68.73_{\pm 0.38}$    & $72.53_{\pm 0.32}$    & $70.63_{\pm 0.34}$   & $63.21_{\pm 0.12}$                       & $62.92_{\pm 0.31}$  \\

FreeSDG \cite{lih}                & $68.57_{\pm 0.19}$    & $71.59_{\pm 0.32}$    & $70.08_{\pm 0.29}$   & $59.72_{\pm 0.13}$                       & $63.53_{\pm 0.19}$  \\
RAS$^{4}$DG(ours)                   & $\textbf{73.52}_{\pm \textbf{0.17}}$   & $\textbf{81.32}_{\pm \textbf{0.14}}$    & $\textbf{76.56}_{\pm \textbf{0.23}}$   & $\textbf{65.86}_{\pm \textbf{0.12}}$                       & $\textbf{72.88}_{\pm \textbf{0.07}}$  \\ \hline
\end{tabular}
\end{table}

\begin{table}[t]
\centering
\caption{Ablation study on different components of the proposed method on the FeTA2021 dataset. The baseline is `w/o SSDG' method.
}
\label{tab:table2}
\newcolumntype{P}[1]{>{\centering\arraybackslash}p{#1}}
\begin{tabular}{p{1.8cm}|P{1.3cm}|P{1.3cm}|P{1.3cm}|P{1.7cm}}

\hline
Methods & RASS & RMS & RSD & DSC \\ \hline
Baseline      &     &    &     & $68.41_{\pm 0.14}$    \\
RASS (only)      & \checkmark    &    &     & $73.23_{\pm 0.19}$    \\
RMS (only)      &      & \checkmark  &     &  $70.92_{\pm 0.12}$   \\
RSD (only)      &      &    & \checkmark   & $69.41_{\pm 0.31}$    \\
w/o RASS      &      & \checkmark  & \checkmark   &  $73.33_{\pm 0.12}$   \\
w/o RMS     & \checkmark    &   & \checkmark    &  $73.56_{\pm 0.18}$   \\
w/o RSD     & \checkmark    & \checkmark  &     & $74.48_{\pm 0.25}$    \\
Ours      & \checkmark    & \checkmark  & \checkmark    & $\textbf{76.56}_{\pm \textbf{0.23}}$    \\ \hline
\end{tabular}
\end{table}

\subsubsection{Ablation Study}
The ablation study is presented in Table \ref{tab:table2}. We adopt the trained baseline model as a lower bound.
These three components improve DSC performance by 4.82\%, 2.51\%, and 1.00\%, respectively.
Specifically, RASS yields the greatest performance improvement by sophisticated synthesis of the amplitude spectrum, while the combination of RMS and RSD better facilitates the learning of structural information. All the components ultimately derive our RAS$^{4}$DG.
The reader is referred to the Supplementary Material for more information on architecture and hyperparametric ablation.

\section{Conclusion}
In this paper, a framework called RAS$^{4}$DG is proposed for performing DG segmentation of medical images from the frequency domain perspective, which includes RASS, RMS, and RSD.
In contrast to standard DG learning schemes, SSDG aims to ensure out-of-domain generalization by using data from only one source dataset. RASS is to introduce perturbations in the amplitude spectrum of medical images to simulate inter-domain variability. We also integrate RMS during training and reconstruction design into the network bottleneck to further improve segmentation performance. Our experimental results on 2D and 3D segmentation tasks show that RAS$^{4}$DG significantly outperforms some recent SSDG methods and DA methods.

\begin{credits}
\subsubsection{\ackname} This work was supported by the Shenzhen Science and Technology Program (JCYJ20230807094104009), in part by the National Natural Science Foundation of China (61873145).

\subsubsection{\discintname}
The authors have no competing interests in the paper.

\end{credits}

%
%
%
%

\end{document}